\documentclass{Interspeech}

\interspeechcameraready

\title{ArVoice: A Multi-Speaker Dataset for Arabic Speech Synthesis}

\author{Hawau Olamide}{Toyin}
\author{Rufael}{Marew}
\author{Humaid}{Alblooshi}
\author{Samar}{M. Magdy}
\author{Hanan}{Aldarmaki}

\affiliation{}{Mohamed Bin Zayed University of Artificial Intelligence}{UAE}
\email{hawau.toyin@mbzuai.ac.ae, hanan.aldarmaki@mbzuai.ac.ae}

\usepackage{comment}
\usepackage{color}
\usepackage{rotating}
\usepackage{tabularray}
\usepackage{subcaption}

\usepackage{booktabs, multicol, multirow}
\newcommand{\ra}[1]{\renewcommand{\arraystretch}{#1}}

\usepackage{pifont}
\newcommand{\cmark}{\ding{51}}
\newcommand{\xmark}{\ding{55}}

\newcommand{\ar}[1]{\<#1>}
\usepackage{arabtex}
\usepackage{utf8}
\usepackage[dvipsnames]{xcolor}

\usepackage[textsize=tiny]{todonotes}
\usepackage{newunicodechar}
\newunicodechar{ː}{:}

\begin{document}

\maketitle

\begin{abstract}
We introduce ArVoice, a multi-speaker Modern Standard Arabic (MSA) speech corpus with diacritized transcriptions, intended  for multi-speaker speech synthesis,  and can be useful for other tasks such as speech-based diacritic restoration, voice conversion, and deepfake detection. ArVoice comprises: (1) a new professionally recorded set from six voice talents with diverse demographics, (2) a modified subset of the Arabic Speech Corpus; and (3) high-quality synthetic speech from two commercial systems. The complete corpus consists of a total of 83.52 hours of speech across 11 voices; around 10 hours consist of human voices from 7 speakers. We train three open-source TTS and two voice conversion systems to illustrate the use cases of the dataset. The corpus is available for research use. 
 

\end{abstract}

\section{Introduction}\label{introduction}

Speech synthesis technologies, such as Text-to-Speech (TTS), and Voice conversion (VC), have shown remarkable progress in quality and naturalness, particularly with the availability of large-scale, high-quality datasets. However, for mid-low resource languages, such as Arabic, the availability of clean and well-curated speech corpora remains limited, posing challenges for developing robust open-source speech synthesis models. Although there are several speech data sets for Arabic ASR \cite{mgb, mubarak-etal-2021-qasr, e1qb-jv46-21}, these sets were mainly sourced from news channels, podcasts, and other sources that are naturally contaminated with noise, overlapping speech, inconsistent recording quality, and expressive or dialectal speech, making them poorly suited for developing robust speech synthesis systems with acceptable quality. Another limitation of these existing resources is the lack of diacritics /taškīl/, which represent important vowel information in Arabic orthography.\footnote{\setcode{utf8} For instance, the following forms of the same root letters /skr/ have three distinct pronunciations and meaning: \text{\scriptsize \<سُكَّرٌ> }/sukːar/ : `sugar',  \text{\scriptsize \< سُكْرٌ> }
 /sukr/ : `Drunkenness',  \text{\scriptsize \<سَكَّرَ>} /sakːara/ : `he closed'.}
 The absence of diacritics results in under-specified text inputs with multiple possible pronunciations, leading to unintelligible synthesized speech. While text-based diacritic restoration models for Arabic could be utilized, these systems were mainly trained on Classical Arabic (CA) text and have been shown to result in high diacritic error rates when applied to Modern Standard Arabic (MSA) speech transcripts \cite{aldarmaki23_interspeech,shatnawi2024automaticrestorationdiacriticsspeech}. 

Several prior works have been published to improve the availability of speech resources for MSA. We group the approaches into two classes: (1) \emph{Recording scenarios}: In this approach, researchers employ voice artists and record them either reading pre-defined texts in carefully set-up environments \cite{kulkarni23_interspeech, cmuartctic} or engaging in spontaneous interview-like conversations \cite{hamed-etal-2020-arzen}, and (2) \emph{Re-purposing}: this approach involves extracting speech from massive online sources like TV recordings, Podcasts, or YouTube \cite{mgb,lin2023voxblinklargescalespeaker}. The first approach results in clean and high-quality speech suitable for training speech synthesis models, but the process is expensive and requires intense manual effort. As a result, most available resources of this kind are limited in size. The second approach exploits available resources, eliminating the need for manual recording, and resulting in large speech corpora, but introduces noise and other sources of variability that often lead to poor synthesis quality.

In this paper, we introduce \texttt{ArVoice}, a multi-speaker  dataset consisting of high-quality speech with fully diacritized transcripts for MSA. \texttt{ArVoice} consists of: (1) professionally recorded audio by two male and two female voice artists from diacritized transcripts, (2) professionally recorded audio by one male and one female voice artists from undiacritized transcripts, (3) repurposed audio from existing single-speaker TTS datasets, and (4) synthesized speech using commercial TTS systems.  The latter is used mainly for voice conversion and TTS data augmentation, and is based on the same text used by human speakers. Overall, the dataset contains $7$ human voices of about $10$ total hours of non-parallel speech, and $8 \times$  $\sim9$ hours of parallel synthesized speech. 

We demonstrate the use of the dataset for TTS and voice conversion. The dataset can also be used for other speech-related tasks, including ASR and speech-based diacritization, making it a valuable resource for the broader research community. The data set accessible to researchers in two forms: the ASC and the synthetic parts of the dataset are available under the Creative Commons Attribution 4.0 International License (CC BY 4.0) through a public link\footnote{\scriptsize\url{https://huggingface.co/datasets/MBZUAI/ArVoice}}. The professionally recorded subsets are granted only to qualified researchers upon signing a formal Data Usage Agreement.

\section{Related Works}
\label{Related Works}

Large and diverse Arabic datasets have been developed for speech tasks; yet, most of them are suitable for ASR, and few are designed for TTS.
The main open-source speech datasets developed specifically for TTS are: ASC and ClArTTS. Both are single-speaker datasets featuring a male speaker. 

\textbf{ASC (Arabic Speech Corpus)} \cite{soton409695} is the most widely known Arabic TTS corpus, and contains around $4$ hours of recorded speech by a single male speaker. Prior to the recording of the corpus, a lot of effort went into curating the text transcripts with complete coverage of the characters and phonemes in the Arabic vocabulary. This was achieved by including many non-sense sentences to maximize the coverage of all phonemic sequences. The realized transcripts for the speech corpus were further modified to reflect the phones, leading to the removal/addition of characters that are silenced/pronounced. In total, the training set has $1.8$K sentences, the test set has $100$ sentences.  \textbf{ClArTTS (Classical Arabic Text-to-Speech)}  \cite{kulkarni23_interspeech} consists of classical Arabic (CA) speech sourced from an open-source digital audiobook. The dataset was developed specifically for end-to-end TTS systems that typically need more data to be effective, resulting in $12$ hours of speech. While it provides consistent and high-quality recordings, it lacks diversity in speakers and Arabic variants, making it less effective for multi-speaker MSA speech synthesis. 
Table \ref{tab:my_label} summarizes the features of these dataset compared to \texttt{ArVoice}. 

\begin{table}[h]
    \centering
    \caption{Comparison of ArVoice with existing Arabic TTS data.}
    \ra{1.1}
    \scalebox{0.85}{%
    \begin{tabular}{@{}lcccc@{}} \toprule
         & ASC & ClArTTS & 
         ArVoice \\ \midrule
        MSA & \cmark & - 
        & 
        \cmark \\
        Number of Human Voices & 1 & 1 & 
        7 \\
        Open-source & \cmark & \cmark 
        & \cmark \\
        Human Speech Duration (hrs) & 4.1 & 12 & 
        10 \\
        Total Duration (hrs) & 4.1 & 12 & 
        83.52 \\
       \bottomrule
    \end{tabular}%
    }
    \label{tab:my_label}
\end{table}

\section{Dataset Construction}

ArVoice comprises both human and synthetic voices. 
In this section, we describe each part of ArVoice and provide justification for design decisions where applicable. 

\subsection{ArVoice Part 1 (Human)}
\label{subsection: ArVoice Part 1}
 \textbf{Text Sources: } 
 Modern Standard Arabic text was sourced from the \textit{Tashkeela} Corpus \cite{tashkeela}, which consists of fully diacritized Arabic text, mostly in Classical Arabic. Within this corpus, we extracted texts from the \textit{Al Jazeera} split, as manual inspection confirmed it to be the only part of the corpus containing MSA. \\

\noindent\textbf{Text Pre-processing:}
Since the \textit{Tashkeela} corpus was originally scraped from the web, it contained numerous markdown elements and extraneous punctuation marks, which we removed as part of preprocessing. The corpus primarily consisted of articles, which we subsequently tokenized into sentences using the \texttt{PyArabic} library\footnote{\url{https://pypi.org/project/PyArabic/}}. Sentences containing English characters or digits (e.g., dates) were initially filtered out. To ensure a realistic audio segment length, sentences exceeding $50$ characters were iteratively split at commas, ensuring that each resulting segment contained at least $20$ characters. After applying these preprocessing steps, we obtained a total of \(\approx1.8K\) diacritized MSA pseudo-sentences; we added $200$ additional pseudo-sentences that included digits for more vocabulary coverage. \\

 \noindent\textbf{Recording:}  
Professional native Arabic voice talents were employed to record the processed text, consisting of two male and two female speakers, ensuring no overlap in recorded texts among the speakers. The talents were fairly compensated and informed about the intended use of their voice. We obtained prior permission for their recordings to be used in Arabic speech synthesis research\footnote{We specified that the requested recordings are intended for research and development of Arabic text-to-speech systems, and that the data may be shared with other researchers for research use only; the license excludes broadcast and resale rights. }.  The speech samples were recorded in quiet rooms using high-quality condenser microphones, with minor post-hoc noise reduction in some instances. \\%

\noindent\textbf{Text Post-Processing:}
The  transcripts were manually inspected to match the actual recordings, as we discovered word repetitions and omissions in some recorded samples. All transcripts were corrected to reflect these repetitions or omissions. For transcripts with digits, the Arabic numerals were first automatically verbalized using the \texttt{PyArabic} library. We carried out manual proofreading and additional annotation mainly to verify diacritic placement and to correct number conversions from digits to words, as the written transcripts often diverged from the pronunciation of the speakers. An example is `2020', which was vocalized as \text{\scriptsize\ar{عِشْرِينْ عِشْرِينْ}} : `twenty twenty', in audio but predicted by PyArabic as  \text{\scriptsize \ar{أَلْفَانِ وَ عِشْرِينْ}} : `two thousand and twenty'.

\subsection{ArVoice Part 2 (Human)}\label{subsection: ArVoice Part 2}

Following the same collection and processing methodology in section \ref{subsection: ArVoice Part 1}, we extracted $0.9$K \textbf{non-diacritized MSA transcripts} from the \textit{Khaleej} Corpus \cite{khaleeji}. The Khaleej Corpus contains news articles on various domains written in MSA, but the text contains no diacritics. \\

\noindent\textbf{Recording:} The Khaleej text was divided into two parts, each recorded by a different voice talent—one male and one female—distinct from the voices in Part 1. Since the text did not contain diacritics, the talents were instructed to read and record it according to what seemed most natural to them. The same prior agreements and recording standards were followed as in Part 1. \\

\subsection{ArVoice Part 3 (Human)}\label{subsection: ArVoice Part 3}

In this part, we utilize the Arabic Speech Corpus (ASC) dataset, which is distributed under the Creative Commons License CC BY 4.0 \footnote{\url{https://en.arabicspeechcorpus.com/} - last accessed on Feb 19, 2025. }. 
The text transcripts in ASC were originally modified to match word pronunciation; for instance, 
\textit{nunation}, which is supposed to be spelled as a diacritic, is instead spelled out as the letter \ar{ن} /n/. Another modification is the omission of \textit{Alef} /a/ before the /l/ in the definite article /al/. These modifications, which were meant to make the dataset better normalized for concatenative TTS systems, are inconvenient for end-to-end systems as they are inconsistent with the standard spelling used in the other parts of the dataset, and require additional pre-processing steps. We used GPT-4 (using prompts proposed in \cite{toyin-etal-2024-sttatts}) to first automatically fix these modifications back to standard spelling, followed by manual correction. Half of the train set of the original ASC corpus \cite{soton409695}, which are duplicates of non-sense sentences (included for phone coverage), were removed completely to obtain 0.9K sentences.
The full test and train set were manually inspected and corrected. \\

\noindent\textbf{Validation:}
We validated the corrections made above by evaluating ASR performance on the test set using three open-source, state-of-the-art Arabic ASR models \cite{radford2022robustspeechrecognitionlargescale,toyin-etal-2023-artst,nemo}. 
Table \ref{tab:zero-shot ASR} shows WER calculated using the original ASC transcripts vs. our (corrected) version; we observe a $\approx 40\% $ drop in Word Error Rate (WER), highlighting the importance of this step for standardizing the transcripts in our set. 

\begin{table}[h]
    \centering
    \ra{1.1}
    \caption{Word error rate (\%) and character error rate (\%) of zero-shot ASR using SOTA models on ArVoice Part 3 and the equivalent subset of ASC. All models are evaluated without diacritics as these models do not generate diacritics at inference.}    
    \resizebox{0.7\columnwidth}{!}{%
    \begin{tabular}{@{}lcc@{}} \toprule
        \textbf{Model} &\textbf{ WER $\downarrow$} & \textbf{CER $\downarrow$} \\ \midrule
        \texttt{ASC}  \cite{soton409695}&& \\
        \hspace{1mm}ArTST \cite{toyin-etal-2023-artst}  & 45.86 & 9.65 \\
        \hspace{1mm}Nvidia CTC-large\cite{nemo} & 45.70 & 8.59 \\
       \hspace{1mm}Whisper-large\cite{radford2022robustspeechrecognitionlargescale}  & 45.22 & 9.94 \\ \midrule
\texttt{ArVoice Part 3} \textbf{(Ours)} && \\
       \hspace{1mm}ArTST \cite{toyin-etal-2023-artst}  & 5.57 & 2.04 \\
       \hspace{1mm}Nvidia CTC-large\cite{nemo} & \textbf{5.10} & \textbf{0.96} \\
       \hspace{1mm}Whisper-large\cite{radford2022robustspeechrecognitionlargescale}  & 7.48 & 2.18 \\
      \bottomrule 
    \end{tabular}%
    }
    \label{tab:zero-shot ASR}
\end{table}

\subsection{ArVoice Part 4 (Synthetic)}
\label{subsection:synthetic}
The final processed transcripts from Parts 1 and 3 were used for synthetic speech generation to create a parallel corpus. We excluded Part 2 as it contains no diacritics. 
Google's commercial TTS models\footnote{\url{https://cloud.google.com/text-to-speech}}  were used to generate varying quality of synthetic MSA speech from all transcripts processed. The audio was generated using both the \textit{Standard} and premium \textit{Wavenet} models for two male and two female voices. We recommend the use of this datasets for data augmentation and voice conversion only as audio quality is not guaranteed.

\subsection{Dataset Statistics}
Table \ref{tab:data} gives a summary of recording information of individual human speakers in \texttt{ArVoice}. This part of \texttt{ArVoice} consists in total of about $9$ hours of training and $0.86$ hours of test data. Although ASC has an average number of words, the speaker has the highest duration as a result of the slower speaking rate \cite{soton409695}. 

\begin{table}[h]
    \centering
    \caption{Data statistics and demographic details per human speaker in \texttt{ArVoice}}
    \ra{1.1}
    \resizebox{\columnwidth}{!}{%
    \begin{tabular}{@{}cccccc@{}} \toprule
        \textbf{Part} &
         \textbf{ID} & \textbf{Gender} &\textbf{Origin} & \# \textbf{ words (K)} & \textbf{Duration (hrs)}   \\ \midrule
        1 & m\_aa & m & Egypt & 14.84 & 1.17   \\
        1& f\_ab & f & Jordan & 17.32 & 1.45  \\
        1& m\_ac & m & Egypt & 21.07 & 1.58  \\
        1& f\_ad & f & Morocco & 15.37 & 1.23   \\
        \midrule
        2& m\_ae & m & Palestine & 6.02 & 0.93   \\
        2& f\_af & f & Egypt & 6.07 & 0.95   \\ 
        \midrule
        3 & m\_asc & m & Syria & 13.21 & 2.69   \\
       \bottomrule
    \end{tabular}%
    }
    \label{tab:data}
\end{table}

\section{Baselines}
In this section, we describe two speech synthesis tasks: multi-speaker text-to-speech (TTS) and voice conversion (VC), trained using \texttt{ArVoice}. 

\subsection{TTS Synthesis Experiments}

\noindent\textbf{Models:} Text-to-Speech synthesis experiments were carried out using three open-source models: \textit{ArTST-tts}~\cite{toyin-etal-2023-artst}, \textit{VITS}~\cite{vits} and \textit{Fish-Speech} \cite{liao-2024}. \textit{ArTST-tts} is an Arabic transformer-based model pre-trained on large amounts of unlabeled speech and subsequently fine-tuned on Classical Arabic data from ClArTTS \cite{kulkarni23_interspeech}. \textit{ArTST-tts} allows for multispeaker training using speaker embeddings, \textit{x-vectors}, extracted using the Speech Brain toolkit\footnote{\scriptsize\url{huggingface.co/speechbrain/spkrec-xvect-voxceleb}}. The model is trained to produce mel spectrograms, and a pre-trained HiFi-GAN vocoder is used to generate the waveform from the generated mel spectrograms. \textit{VITS} (Variational Inference Text-to-Speech) integrates variational autoencoders, normalizing flows, and adversarial training. It is an end-to-end architecture that uses a monotonic alignment search mechanism. \textit{Fish-Speech} employs a serial fast-slow Dual Autoregressive (Dual-AR) architecture to enhance the stability of Grouped Finite Scalar Vector Quantization (GFSQ) in sequence generation tasks. Additionally, the system leverages LLMs for linguistic feature extraction and a new neural vocoder architecture, FF-GAN, achieving superior compression ratios. \\

\noindent\textbf{Experimental Settings:} \label{sec: tts exp setting}
 VITS and Fish-Speech originally did not support Arabic TTS, but ArTST-tts was already trained on non-diacritized Arabic text. For all models, we fine-tuned a multi-speaker checkpoint using Parts 1,2,3 of ArVoice (the human voices with diacritized text). For fine-tuning, we used the default hyperparameter settings as predefined for each model and their pre-trained checkpoints.\footnote{%
     \scriptsize
  \begin{tabular}[t]{@{}l@{}}
\url{https://github.com/jaywalnut310/vits}\\
    \url{https://github.com/mbzuai‐nlp/ArTST}\\
    \url{https://github.com/fishaudio/fish‐speech}
  \end{tabular}%
}
For VITS, slight modifications were made to its text cleaner functions to handle Arabic text. \\

\noindent\textbf{Evaluation Methodology:} 
\label{sec: results}
We first compare the performance of the three TTS models in two variants: with/without diacritics in text. We conducted pairwise subjective preference tests, where we randomly sampled $25$ transcripts from the test set and presented synthesized speech from the two model variants. We used the Prolific\footnote{\scriptsize\url{https://app.prolific.com/}} crowd-sourcing platform to employ native Arabic speakers for the preference test, who were presented with two synthesized samples of the same underlying text from systems A and B, and asked to select their preference: A, B, or No preference.
For each preference test, a minimum of $10$ different evaluators rated each of the 25 audio pairs. We aggregated the ratings per sample and calculated the mean and standard error for each category: A, B, and No Preference, then estimated the 95\% confidence interval using the t-distribution. \\

\noindent\textbf{Evaluation Results:}
Figure 1 presents the results for the three TTS models, where pairwise preference tests were conducted to compare the versions trained with and without diacritics for each model. The only model with statistically significant preference is VITS, where the version with diacritics was clearly preferred over the one without. For ArTST and Fish Speech, although a slight preference for the version without diacritics was observed, the difference was not statistically significant. 

Following these results, we further evaluated the intelligibility of the synthesized speech using an ASR model as an automatic metric for sanity check. The results are presented in Table \ref{tab:intelligibility}. The results indicate that speech synthesized by Fish-Speech is of extremely low quality, as WER is above 100\%. Subjective listening tests also confirmed the low intelligibility of this model. We exclude this model from further analysis. 

As VITS with diacritics performed on a par with ArTST in the intelligibility metric, but ArTST without diacritics was preferred in the previous subjective test, we performed another subjective test comparing VITS with diacritics against ArTST without diacritics. The results show a strong and statistically significant preference for VITS (See Figure 2 (a)), which indicates that VITS is the best among the tested three models on this dataset. Finally, we compared the performance of VITS in multi-speaker vs. single-speaker settings (Figure 2 (b)) using the data for m\_asc speaker with $2.69$ hours of speech for training the single-speaker model. Both models are evaluated on the same speaker. A strong preference for the multi-speaker model is observed, illustrating the advantage of more data in training TTS systems even for improving the same target voice. 

\begin{table}[h]
    \centering
    \caption{Intelligibility evaluation using WER(\%) of ASR predictions. We use the ASR model from \cite{toyin-etal-2023-artst} to obtain WER}
    \ra{1.2}
    \resizebox{0.7\columnwidth}{!}{%
    \begin{tabular}{@{}lccc@{}} \toprule
         & \textbf{ArTST-tts} & \textbf{Fish-Speech} & \textbf{VITS} \\ \midrule
      \emph{with diacritics} &  &  \\
       \checkmark & 21.81 & 162.04 & \textbf{20.67}\\
       \xmark & 21.24  & 115.58 & 35.69 \\ \bottomrule
    \end{tabular}%
    }
    \label{tab:intelligibility}
\end{table}

\begin{figure}[t]
\centering  \includegraphics[width=0.7\linewidth]{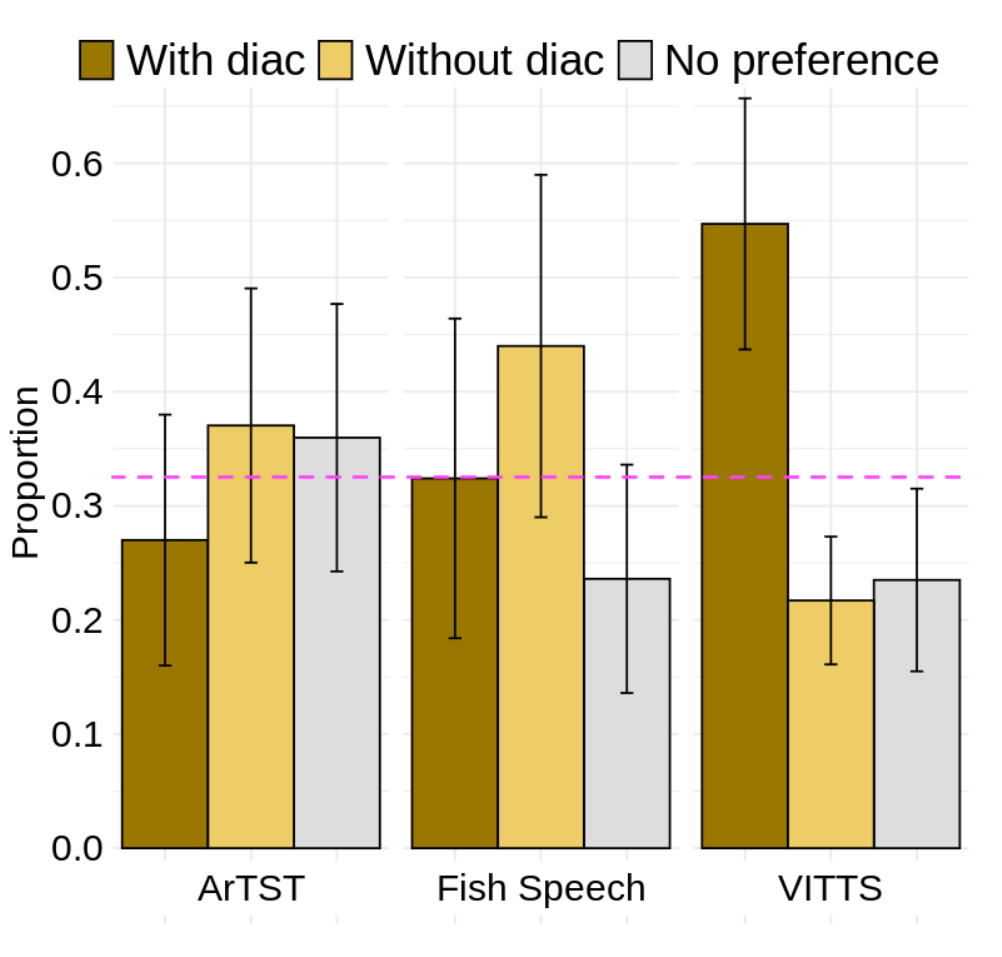}
    \caption{Mean average preference with/without diacritics for each TTS model, with 95\% confidence intervals.}
    \label{fig:diac_err}
\end{figure}

\begin{figure}
\centering
  \subcaptionbox{VITS (w. diac) vs. ArTST (w.o. diac) \label{fig3:a}}{\includegraphics[width=1.2in]{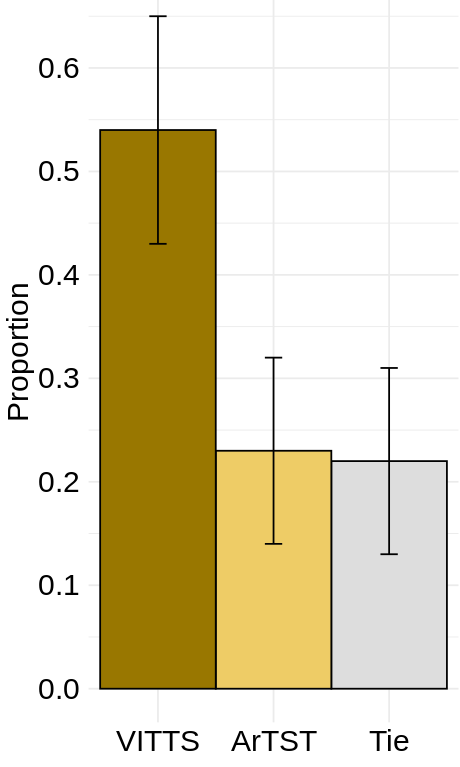}}\hspace{1em}%
  \subcaptionbox{Mutli-Speaker vs. Single-Speaker VITS (w. diac)\label{fig3:b}}{\includegraphics[width=1.2in]{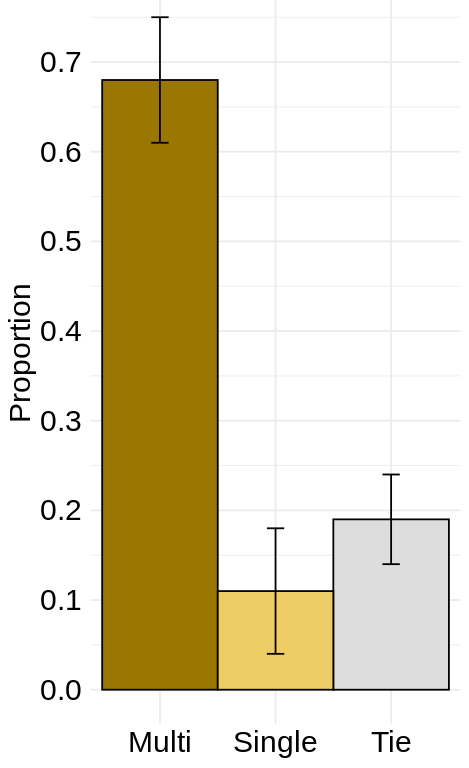}}
\caption{Mean average preference with 95\% confidence intervals. We compare (a) the best version of VITS against. the best version of ArTST, and (b) Multi-speaker vs. single-speaker TTS with the VITS model.}  
\end{figure}

\noindent\textbf{Effect of Synthetic Data Augmentation:}
We trained VITS on the full ArVoice corpus (synthetic and human speech) for half the number of epochs as in section \ref{sec: tts exp setting}; the intelligibility of the synthesized speech improved with an 8.2\% drop in absolute WER compared to the version trained on human speech alone. Subjective preference tests also indicated a preference for models with synthetic data augmentation (about 44\% of raters preferred the version with data augmentation, compared with 30\% and 26\% for the baseline and No Preference, respectively).

\subsection{Voice Conversion (VC) Models}

We experimented with AAS-VC \cite{huang2023aasvc}, a parallel VC model based on a non-autoregressive sequence-to-sequence architecture. To train this model, we utilized 4 speakers from Part 4 of ArVoice (see Section \ref{subsection:synthetic}), which consists of synthetic parallel speech. We trained 4 parallel VC models and averaged their performance. We also fine-tuned a ParallelWaveGAN vocoder \cite{yamamoto2020parallel} on the same dataset to improve waveform synthesis quality. We also fine-tuned KNN-VC \cite{baas2023voice}, a non-parallel VC model that converts source into target speech by replacing each frame of the source representation with its nearest neighbor from the target. To train this model, we used human voices in Arvoice Parts 1 (section \ref{subsection: ArVoice Part 1}) and 3 (section \ref{subsection: ArVoice Part 3}). 
All the fine-tuning processes were conducted using the default parameters specified in the original studies\footnote{%
  \scriptsize
  \begin{tabular}[t]{@{}l@{}}
    \url{https://github.com/bshall/knn-vc}\\
    \url{https://github.com/rufaelfekadu/seq2seq-vc}
  \end{tabular}%
}.
To evaluate the quality of the converted speech, we used the pre-trained  ECAPA-TDNN speaker verification system from SpeechBrain\footnote{\scriptsize
\url{huggingface.co/speechbrain/spkrec-ecapa-voxceleb}} to test whether the converted speech is indistinguishable from real speech. We report the average Speaker Similarity (SS) scores between the target and generated audios; we also report False Acceptance Rate (FAR) at a 0.5 similarity threshold. FAR is correlated with SS. 

\noindent\textbf{Results:}
The KNN-VC model achieved a FAR of \textbf{0.81} with an average SS score of 0.69, indicating fair similarity between the converted voice and the target speaker. The AAS-VC model achieved a higher FAR of \textbf{0.95} with an average SS score of 0.72, indicating relatively high similarity\footnote{The ground truth SS score was 0.81.}. 


\section{Conclusion}
We described ArVoice, a multi-speaker dataset for speech synthesis in Modern Standard Arabic. The dataset consists of $11$ voices in total, 7 of which are human voices, and 4 are synthetic with parallel text. We illustrated the usability of ArVoice in multi-speaker TTS, demonstrating the advantage of using diacritized transcripts. In addition, we demonstrated the benefit of the synthetic speech for TTS data augmentation, as well as parallel voice conversion. ArVoice is the largest open-source Arabic dataset curated specifically for speech synthesis, both in terms of speech duration (83.5 hrs) and number of voices (11).

\section{Acknowledgments}
\ifinterspeechfinal
We thank the five hired voice artists and our co-author, Samar Magdy, for contributing their voice to research. This work was partially funded by a Google research award (11/2023).


\bibliographystyle{IEEEtran}
\bibliography{mybib}

\end{document}